\documentclass[runningheads]{llncs}
\usepackage[T1]{fontenc}
\usepackage{graphicx,verbatim}
\usepackage{marvosym}
\DeclareRobustCommand{\cormark}{\textsuperscript{(\Letter)}}

\usepackage[dvipsnames]{xcolor}
\usepackage{hyperref}
  \hypersetup{
      colorlinks=true,
      urlcolor=blue
  }

\usepackage{booktabs}
\usepackage{multirow}
\usepackage{amsfonts}
\usepackage{amssymb}
\usepackage{capt-of}
\begin{document}
\title{SonoCLIP: Mask-Guided Region-Aware Vision–Language Pretraining for \\ Fetal Ultrasound Analysis}
%
\titlerunning{SonoCLIP}

\author{
Hang Su\inst{1}\thanks{Equal contribution.} \and
Chao Sun\inst{1,2}\protect\footnotemark[1] \and
Zhaofan Li\inst{1} \and
Wei Hu\inst{3}\\[0.35em]
Juhua Liu\inst{1,2,4,5}\cormark \and
Bo Du\inst{1,2,4,5}\cormark
}

\authorrunning{H. Su et al.}

\institute{
School of Computer Science, Wuhan University,
Wuhan, China \and
Institute of Artificial Intelligence, Wuhan University,
Wuhan, China \and
Department of Ultrasound, Renmin Hospital of Wuhan University,
Wuhan, China \and
National Engineering Research Center for Multimedia Software,
Wuhan University, Wuhan, China \and
Hubei Key Laboratory of Multimedia and Network Communication
Engineering, Wuhan University, Wuhan, China\\
\email{\{liujuhua@whu.edu.cn, dubo@whu.edu.cn\}}
}

\maketitle              
\begin{abstract}

Vision–language foundation models have shown strong potential in medical image analysis. Although foundation models for ultrasound imaging have recently emerged, the domain is particularly challenging due to severe speckle noise, acquisition variability, and subtle anatomical boundaries, leading to high inter-observer variability. Existing CLIP-based models rely primarily on global image–text alignment, limiting their sensitivity to clinically decisive local structures. We propose SonoCLIP, the first million-scale region-controllable fetal ultrasound vision-language foundation model that integrates segmentation masks as mask-channel visual prompts within an encoder, enabling joint global–local contrastive representation learning. To support scalable region–text alignment, we introduce a sigmoid-based pairwise contrastive loss that improves stability under large-scale supervision. We further curate a 1.44M-image multimodal fetal ultrasound dataset spanning 24 standard planes for large-scale pretraining. Extensive cross-center evaluations demonstrate that SonoCLIP achieves superior zero-shot transfer performance under both global and mask-guided inference, establishing a controllable and clinically oriented foundation model for fetal ultrasound analysis. Our code and data are
available at:
  \href{https://github.com/Harrison-one/SonoCLIP}{https://github.com/Harrison-one/SonoCLIP}.


\end{abstract}
%
%
\section{Introduction}
\label{sec:intro}

Prenatal care has been significantly advanced by ultrasound technology. Due to its safety, accessibility, and cost-effectiveness, fetal ultrasound is routinely used for real-time monitoring of fetal development and early detection of congenital abnormalities \cite{10.1145/3447243,2024ISUOG,2013ISUOG}. However, ultrasound interpretation remains subjective and operator-dependent, leading to substantial inter-observer variability. Image quality is often degraded by speckle noise and view-dependent artifacts. As a result, less-experienced clinicians may fail to identify subtle anatomical structures \cite{Singh2025ExplainableFetalUNetpp,Guo2026SonomateFetalUltrasound}. This limitation causes inconsistent fetal biometry measurements and hinders standardized assessment in clinical practice \cite{BurgosArtizzu2020MaternalFetalPlaneCNN}. The problem is more severe in resource-limited settings, where experienced sonographers are scarce. These challenges highlight the urgent need for robust and trustworthy artificial intelligence methods to enhance diagnostic consistency and reliability in fetal ultrasound imaging \cite{Krishna2024StackedEnsembleFetalPlanes,He2024FCUMNeurocomputing,Yeung2024ImplicitVol}.

In recent years, foundation models (FMs) trained with self-supervised learning have achieved notable success and advanced medical artificial intelligence \cite{Ma2025ArkPlusCXR,Ma2025GPFM,Yan2025PanDerm}. By learning generic representations from large-scale unlabeled data, FMs can be adapted to diverse downstream tasks with minimal annotation. This paradigm outperforms traditional task-specific models that require extensive retraining. Vision–language models such as CLIP align visual and textual features through contrastive learning and demonstrate strong transferability \cite{Radford2021CLIP,10847310}. Recent extensions to biomedicine further validate this potential \cite{Zhang2023BiomedCLIP,Khattak2024UniMedCLIP,Maani2025FetalCLIP}. However, a widely accepted foundation model for fetal ultrasound is still lacking. 
Existing approaches face several limitations. First, most methods adopt generic visual backbones \cite{He2022MAE} and rely on relatively small ultrasound datasets, which constrain robust fine-tuning. Models such as CLIP, pretrained on natural images, lack domain-specific anatomical knowledge. Although UniMed-CLIP \cite{Khattak2024UniMedCLIP} leverages large-scale multimodal medical data, its modality-specific performance remains inconsistent and underexplored for fetal ultrasound.
Second, fetal view recognition and cross-center generalization require sensitivity to subtle local structures, such as small chambers and fine boundaries. These cues are often degraded by speckle noise and acquisition variability. Global image–text alignment captures coarse semantics but may overlook clinically critical details. Moreover, FetalCLIP \cite{Maani2025FetalCLIP} does not explicitly address the localization–context trade-off in fetal imaging, nor does it incorporate architectural designs tailored to the spatial hierarchy and morphological complexity of fetal anatomy.

We propose \textbf{SonoCLIP}, a foundation model for fetal ultrasound that introduces explicit region controllability during large-scale contrastive fine-tuning. Segmentation annotations are treated as visual prompts. Specifically, masks are appended as an additional channel to the image encoder, enabling region-guided representation learning while preserving global context. To effectively leverage both global- and region-level image–text pairs, we replace the conventional softmax-based contrastive objective with a sigmoid pairwise alignment loss. This formulation optimizes each pair independently and reduces sensitivity to batch composition. SonoCLIP is trained on a million-scale dataset covering 24 standard fetal ultrasound planes. We evaluate cross-center zero-shot plane classification under both global and mask-guided inference. 

Our main contributions are summarized as follows:

\begin{itemize}
\item We propose a region-controllable fetal ultrasound foundation model, SonoCLIP, that incorporates mask-channel prompts to enable region-aware contrastive representation learning.

\item We introduce a sigmoid-based pairwise contrastive loss to enhance stability and scalability for large-scale region–text fine-tuning.

\item We curate and utilize a \textbf{1.44M}-image multimodal fetal ultrasound dataset with plane labels, segmentation annotations, and structured captions, and demonstrate strong cross-center zero-shot transfer on \textbf{24} plane classes with both global and mask-guided inference.
\end{itemize}

\section{Method}

\begin{figure}[t]
    \centering
    \includegraphics[width=1.0\linewidth]{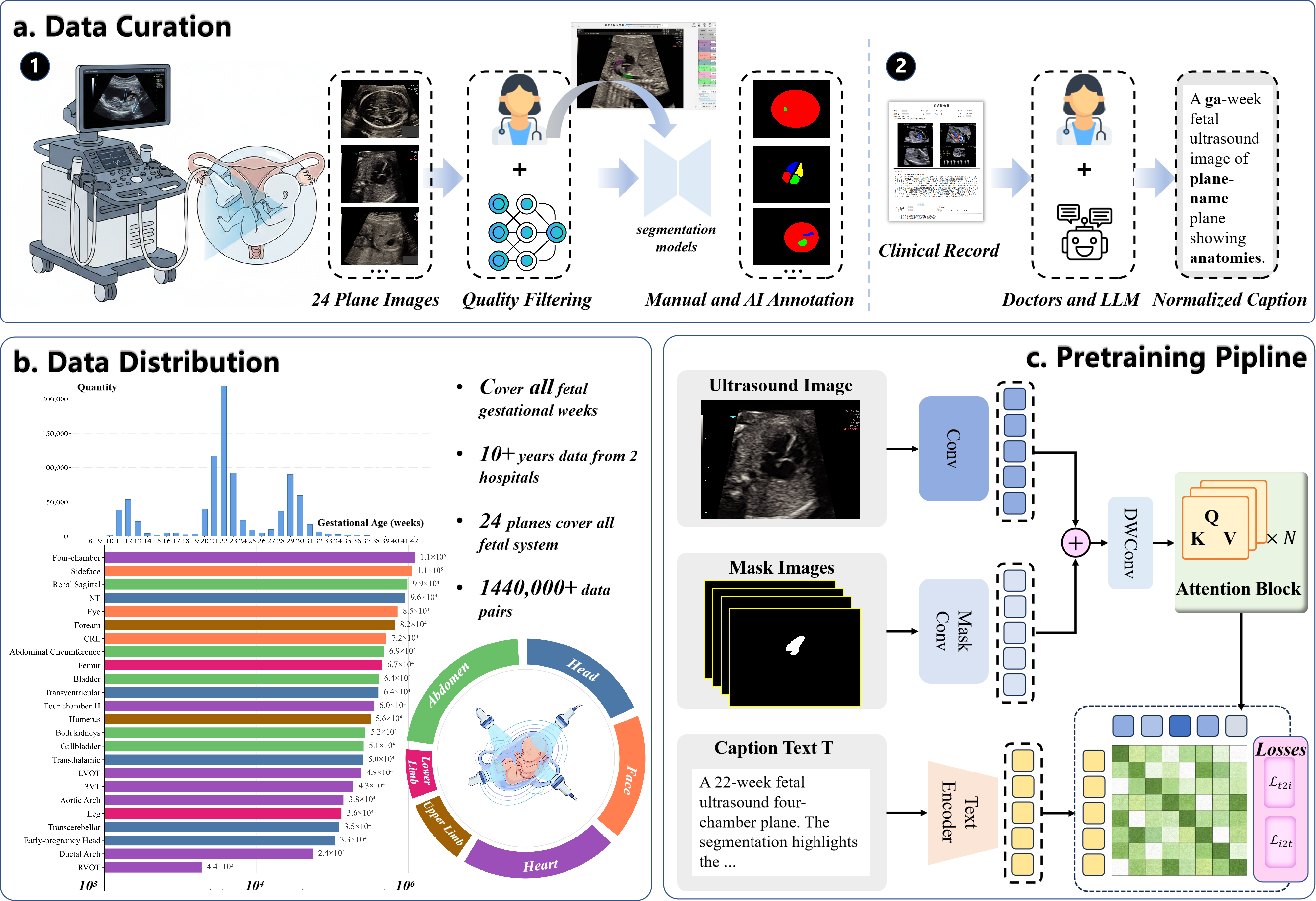}
    \caption{Overview of SonoCLIP. (a) Data curation pipeline, including quality control, manual and AI-assisted mask annotation, and standardized caption generation from clinical records. (b) Dataset distribution across gestational weeks, hospitals, and anatomical planes. (c) Pre-training framework: ultrasound images and masks are jointly encoded via a mask-channel visual pathway, and aligned with paired text using a sigmoid-based objective.}
    \label{fig:fig1}
\end{figure}

\subsection{Data Curation and Region--Text Pair Generation}

\subsubsection{AI-assisted Mask Extraction and Curation} 
As illustrated in Fig.~\ref{fig:fig1}(a), we first manually annotated 500 images for each anatomical plane to train an initial segmentation model. This model was then applied to each ultrasound image to generate candidate anatomical masks with predicted labels and confidence scores. To address fragmented predictions or duplicate instances under the same label, we perform label-wise consolidation. For multiple disconnected regions with identical labels, only the largest connected component is retained. We further filter masks using confidence thresholds and basic geometric constraints, such as non-empty foreground and plausible spatial extent, to remove unreliable predictions. This procedure produces a curated set of structural masks for each image, reducing noise in region-level supervision.

\subsubsection{Region–Text Pair Generation for Fetal Ultrasound}
Following mask filtering, we construct region-text pairs for training. For each filtered image-mask pair, we employ a large language model and templates to generate structure-specific descriptions, exemplified by:
\begin{quote}
\textit{A \textbf{ga}-week fetal ultrasound \textbf{plane-name} plane. The segmentation highlights the \textbf{anatomy}.}
\end{quote}
To retain plane-level semantics, we generate a global description for each image:
\begin{quote}
\textit{A \textbf{ga}-week fetal ultrasound image of the \textbf{plane-name} plane showing \textbf{anatomies}.}
\end{quote}
During training, global image–text pairs are combined with mask-based region–text pairs. This strategy enables joint learning of plane-level recognition and anatomically grounded alignment at scale.

\subsection{Mask-channel Visual Pathway}

Inspired by Alpha-CLIP \cite{Sun_2024_CVPR}, we introduce a mask-channel visual pathway to jointly encode each ultrasound image and its corresponding anatomical mask. This design injects region-specific cues into the visual encoder while retaining the global contextual representations of the original CLIP image branch. Given a pseudo-color ultrasound image $\mathbf{X}\in\mathbb{R}^{H\times W\times 3}$ and a binary anatomical mask $\textbf{A}=\textbf{M}\in\mathbb\{0,1\}^{H\times W}$, where 1 denotes the foreground and 0 the background, we process them through two parallel convolutional stems and fuse their features at the embedding level:

\begin{equation}
\textbf{E}=\mathrm{DWConv}(\mathrm{Conv}(\textbf{X})+\mathrm{Conv}_{m}(\textbf{A})),
\end{equation}
where $\mathrm{DWConv}$ denotes depthwise convolution, $\mathrm{Conv}$ is the original image convolution, and $\mathrm{Conv}_{m}$ is the mask-channel convolution with the same kernel size and stride. As a result, both branches produce embeddings with identical spatial resolution and feature dimension.

To preserve the pretrained CLIP behavior at initialization, we initialize the Mask Conv kernel weights to zero,
which ensures $\mathrm{Conv}_{m}(\textbf{A})=0$ initially and thus $\textbf{E}=\mathrm{Conv}(\textbf{X})$. During fine-tuning, $\mathrm{Conv}_{m}$ learns to inject spatially localised cues into the early representations based on mask prompts, thereby achieving region-specific encoding while preserving the prior information embedded in the pre-trained original path.

\subsection{Sigmoid Pairwise Contrastive Loss}

Our goal is to learn a joint embedding space where prompted ultrasound representations align with their matched texts at both global and structural levels, while remaining distinct from unmatched texts within the same mini-batch. The supervision is heterogeneous, combining global captions and mask-based region descriptions. In addition, the number of implicit negatives increases with batch size. Under this setting, softmax-normalized objectives such as InfoNCE couple all samples through batch-level normalization and are sensitive to batch composition. Inspired by SigLIP \cite{Zhai_2023_ICCV}, we adopt a pairwise sigmoid alignment loss. Each image–text pair is optimized independently as a binary matching objective, while still benefiting from large numbers of implicit negatives.

Let $v_i$ and $p_i$ be L2-normalized image and text embeddings:
\begin{equation}
v_i=\frac{f_\theta(\hat{x}_i)}{\|f_\theta(\hat{x}_i)\|},\quad p_j=\frac{g(\hat{l}_j)}{\|g(\hat{l}_j)\|}.
\end{equation}

We compute pairwise logits:
\begin{equation}
s_{ij}=t\cdot\langle v_i,t_j\rangle+b,
\end{equation}
where $t$ is a learnable logit scale and $b$ is a learnable bias. For a matched pair we set $y_{ij}=+1$, otherwise $y_{ij}=-1.$ The image-to-text loss is defined as:
\begin{equation}
\mathcal{L}_{i2t}=\frac1n\sum_i\sum_j\log(1+\exp(-y_{ij}s_{ij}))\:,
\end{equation}
and we analogously compute a text-to-image loss $\mathcal{L}_{t2i}$ by swapping the roles of image and text features. Final alignment loss is the symmetric average:

\begin{equation}
\mathcal{L}_{sig}=\frac{1}{2}\left(\mathcal{L}_{i2t}+\mathcal{L}_{t2i}\right).
\end{equation}

\section{Experiments}

\subsection{Datasets and Experimental Details}

\subsubsection{FetalP24 Dataset}:
FetalP24 dataset(Center A) contains 1.44 million fetal ultrasound images, covering 24 standard anatomical planes. The dataset currently includes 36,482 patients, mainly acquired from GE Voluson E8/E10 systems. Each image is annotated with anatomical masks and gestational age (GA) labels. The plane-mask correspondences are defined as follows: Abdominal Circumference (Abdominal Circumference, Stomach, Spine), Aortic Arch (Aortic Arch, Ascending Aorta, Descending Aorta, Aortic Isthmus), Bladder (Bladder), Both Kidneys (Kidneys, Renal Pelvis), Both Orbits (Orbits), Ductal Arch (Ductal Arch), Early-pregnancy Head (Early-pregnancy Head Circumference), Femur (Femur), Forearm (Forearm, Ulna, Radius), Four-chamber (Left Atrium, Right Atrium, Left Ventricle, Right Ventricle), Four-chamber Horizontal (Left Ventricular Wall, Right Ventricular Wall), Gallbladder (Gallbladder), Humerus (Humerus), Left Ventricular Outflow Tract (Left Ventricular Outflow Tract), Leg (Leg, Tibia, Fibula), Mid-sagittal (Crown-rump), Nuchal Translucency (Nuchal Translucency, NT Nasal Bone), Renal Sagittal (Kidney), Right Ventricular Outflow Tract (Right Ventricular Outflow Tract), Side Face (Nasal Bone), Three-vessel and Trachea (Aortic Arch, Ductus Arteriosus), Transcerebellar (Cerebellum, Cisterna Magna, Transcerebellar Head Circumference), Transthalamic (Transthalamic Head Circumference, Cavum Septi Pellucidi), and Transventricular (Lateral Ventricle, Choroid Plexus, Transventricular Head Circumference).Textual descriptions of both global and structure-specific features were generated using templates. A FetalP24(Center B) test dataset  comprises the same 24 categories, with approximately 100 images per category, for zero-shot evaluation under both global and mask-guided inference. Images from Centre B were never utilised during training. We publicly release a feature-level version of the test set.

\subsubsection{FetalP6 Dataset}:
We further constructed a FetalP6 dataset comprising 5034 images by integrating two publicly available benchmark datasets: MFP\cite{ashkani2022fast} and FETAL-PLANES-DB\cite{burgos2020evaluation}. Additionally, we refined the HC18 subset within MFP into early pregnancy head and corpus callosum plane categories. The final dataset comprises six standard planes: Abdominal Circumference(AC), Cerebellum(Cereb), Early-pregnancy Head(EarlyHead), Femur, Four-chamber, and Transthalamic.  Concurrently, we curated a segmentation
FetalP5 dataset comprising five plane-mask pairs: Abdominal Circumference (Abdominal Circumference), Cerebellum (Cerebellum), Early-pregnancy Head (Early-pregnancy Head Circumference), Femur (Femur), and Transthalamic (Transthalamic Head Circumference).

\subsubsection{Experimental Setup}
For SonoCLIP pre-training, we employed CLIP (ViT-L/14@336px) as the foundational architecture, training on four NVIDIA RTX 3090 GPUs with an effective batch size of 32. We employed the AdamW algorithm with a base learning rate of $1\times10^{-4}$ and a weight decay rate of 0.02. The masking branch was optimised using the full learning rate, while the remaining visual encoder parameters, learnable log probability scaling factors, and biases were trained with a learning rate multiplier of 0.01×. The learning rate was scheduled via 5000 warm-up steps followed by cosine decay. To maintain global semantic alignment, 10\% of training pairs employed all-ones masks and full descriptions. 
For downstream evaluation, we adopt a frozen-encoder protocol for both classification and segmentation. For classification, we train only a lightweight linear head. For segmentation, we optimize a lightweight decoder on top of dense patch-level features. To prevent label leakage, segmentation uses an all-ones mask-channel input rather than the ground-truth mask.

\begin{center}
\begin{minipage}{0.46\textwidth}
\centering
\captionof{table}{Zero-shot classification performance on the FetalP24 dataset (Center B). Best results are in \textbf{bold}; second-best are \underline{underlined}.}
\label{tab:zeroshot_results}
\small
\setlength{\tabcolsep}{4.5pt}
\begin{tabular}{lcc}
\toprule
\textbf{Method} & \textbf{Top-1} & \textbf{Top-5} \\
\midrule
CLIP & 10.52 & 29.59 \\
UniMed-CLIP & 16.69 & 50.34 \\
FetalCLIP & 39.78 & 83.25 \\
\midrule
SonoCLIP (w/o mask) & \underline{58.38} & \underline{94.47} \\
SonoCLIP (w/ mask) & \textbf{85.01} & \textbf{99.01} \\
\bottomrule
\end{tabular}
\end{minipage}
\hfill
\begin{minipage}{0.48\textwidth}
\centering
\includegraphics[width=\linewidth]{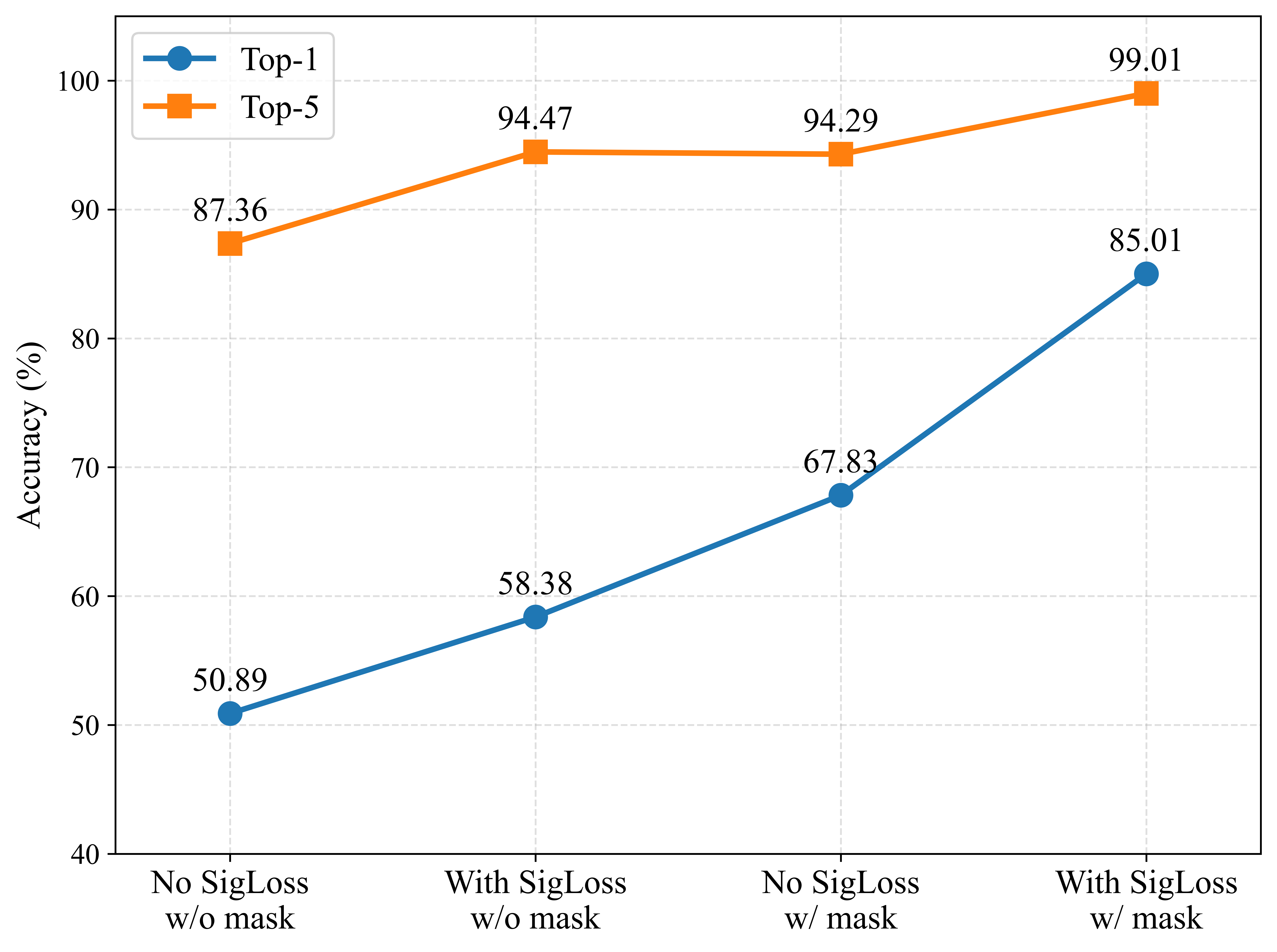}
\captionof{figure}{Ablation study of SigLoss and mask guidance in zero-shot classification on the FetalP24 dataset.}
\label{fig:ablation_sigloss_mask}
\end{minipage}
\end{center}

\subsection{Zero-shot Classification on the FetalP24 Dataset}
As shown in Table~\ref{tab:zeroshot_results}, CLIP and UniMed-CLIP achieve relatively low performance, indicating limited transferability to fetal ultrasound. FetalCLIP performs substantially better, highlighting the importance of domain adaptation. SonoCLIP (w/o mask) further improves over FetalCLIP by 18.60 and 11.22 percentage points in Top-1 and Top-5 accuracy, respectively, showing that the proposed training framework learns more transferable representations under mixed global and region-level supervision. With mask-guided inference, SonoCLIP (w/ mask) gains a further 26.63\% and 4.54\% over SonoCLIP (w/o mask), indicating that mask guidance provides additional benefits by emphasizing clinically relevant anatomical regions during zero-shot matching.

\subsection{Ablation Studies}
Fig.~\ref{fig:ablation_sigloss_mask} presents ablation results for SigLoss and mask-guided learning in zero-shot classification. When mask-guided learning is disabled, incorporating SigLoss alone yields a modest improvement in model performance. However, enabling both mask-guided learning and SigLoss simultaneously results in a significant increase in both Top-1 and Top-5 performance. This indicates that SigLoss's advantages become more pronounced when combined with mask guidance, and also demonstrates a strong synergistic effect between paired alignment objectives and mask-assisted inference.

\begin{figure}[t]
    \centering
    \includegraphics[width=0.99\linewidth]{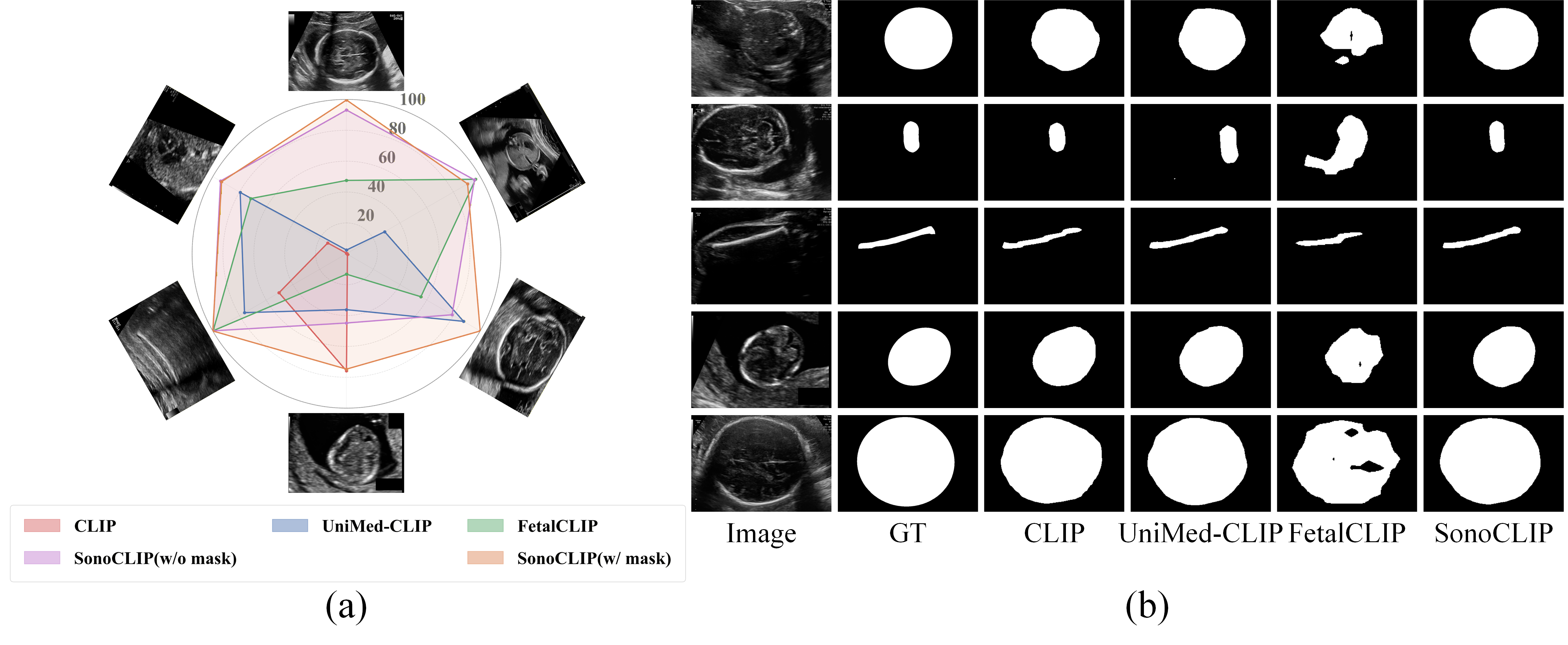}
    \caption{(a) Zero-shot classification results of each model on the FetalP6 dataset. (b) Qualitative comparison of segmentation results on the FetalP5 dataset.}
    \label{fig:zero_shot__seg_open_data}
\end{figure}
\subsection{Zero-shot Classification on the FetalP6 Dataset}

Fig.~\ref{fig:zero_shot__seg_open_data}(a) reports zero-shot classification on the processed FetalP6 dataset, with no training or fine-tuning on this dataset. SonoCLIP without masks (w/o mask) performs strongly across representative planes and outperforms baseline models in overall balance, indicating improved cross-dataset generalization. Mask-assisted (w/ mask) inference further boosts accuracy on several planes, suggesting additional gains from region guidance.

\begin{table}[h]
\centering
\caption{Linear-probe classification performance on the FetalP6 dataset.``w/'' and ``w/o'' denote with and without the mask-channel pathway, respectively. Best results are highlighted in \textbf{bold}, and second-best results are \underline{underlined}.}
\label{tab:sixplane_cls}
\setlength{\tabcolsep}{3pt}
\resizebox{\textwidth}{!}{%
\begin{tabular}{lcccccccccccccc}
\toprule
\textbf{Model}
& \multicolumn{2}{c}{\textbf{AC}}
& \multicolumn{2}{c}{\textbf{Cereb}}
& \multicolumn{2}{c}{\textbf{Earlyhead}}
& \multicolumn{2}{c}{\textbf{Femur}}
& \multicolumn{2}{c}{\textbf{Four-chamber}}
& \multicolumn{2}{c}{\textbf{Transthalamic}}
& \multicolumn{2}{c}{\textbf{Avg}} \\
\cmidrule(lr){2-3} \cmidrule(lr){4-5} \cmidrule(lr){6-7}
\cmidrule(lr){8-9} \cmidrule(lr){10-11} \cmidrule(lr){12-13}
\cmidrule(lr){14-15}
& Acc & F1
& Acc & F1
& Acc & F1
& Acc & F1
& Acc & F1
& Acc & F1
& Acc & F1 \\
\midrule
CLIP
& 89.3 & 87.4
& 50.7 & 58.0
& 91.4 & \underline{95.5}
& 97.6 & 96.2
& 90.7 & 92.6
& 89.7 & 84.9
& 87.1 & 85.8 \\
UniMed-CLIP
& 91.6 & 87.7
& 69.1 & 57.7
& \textbf{100.0} & 87.5
& \underline{98.6} & 97.9
& 88.1 & 91.1
& 66.9 & 74.6
& 83.5 & 82.7 \\
FetalCLIP
& \underline{97.3} & 97.8
& 80.1 & 80.7
& 77.1 & 85.7
& \textbf{100.0} & 99.3
& \underline{99.1} & 99.1
& 92.3 & 91.1
& 94.6 & 92.3 \\
\midrule
SonoCLIP(w/o)
& \textbf{98.2} & \textbf{98.9}
& \underline{83.1} & \underline{85.0}
& \underline{94.3} & \textbf{95.7}
& \textbf{100.0} & \underline{99.5}
& \textbf{100.0} & \underline{99.9}
& \underline{94.2} & \underline{93.2}
& \underline{96.3} & \underline{95.3} \\
SonoCLIP(w/)
& \textbf{98.2} & \underline{98.7}
& \textbf{100.0} & \textbf{100.0}
& 91.4 & \underline{95.5}
& \textbf{100.0} & \textbf{99.8}
& \textbf{100.0} & \textbf{100.0}
& \textbf{99.4} & \textbf{98.7}
& \textbf{99.3} & \textbf{98.8} \\
\bottomrule
\end{tabular}%
}
\end{table}

\subsection{Linear Probe Classification on the FetalP6 Dataset}

Table~\ref{tab:sixplane_cls} reports linear-probe classification results. SonoCLIP achieves the best overall performance among all methods, improving mean accuracy and F1 by $3.0\%$ and $3.5\%$ over FetalCLIP, respectively. Notably, this method achieves top results across most planes, demonstrating consistent robust classification capabilities across multi-class fetal ultrasound planes. This also indicates that the proposed pre-training strategy can learn more discriminative and robust fetal ultrasound representations, while mask-guided techniques further enhance downstream classification performance.

\begin{table}[h]
\centering
\caption{Segmentation performance on the FetalP5 dataset. ``w/o'' denotes without the mask-channel pathway. Best results are highlighted in \textbf{bold}, and second-best results are \underline{underlined}.}
\label{tab:five_plane_seg}
\setlength{\tabcolsep}{3.5pt}
\resizebox{\textwidth}{!}{%
\begin{tabular}{lcccccccccccccc}
\toprule
\textbf{Model}
& \multicolumn{2}{c}{\textbf{AC}}
& \multicolumn{2}{c}{\textbf{Cereb}}
& \multicolumn{2}{c}{\textbf{Earlyhead}}
& \multicolumn{2}{c}{\textbf{Femur}}
& \multicolumn{2}{c}{\textbf{Transthalamic}}
& \multicolumn{2}{c}{\textbf{Avg}} \\
\cmidrule(lr){2-3} \cmidrule(lr){4-5} \cmidrule(lr){6-7}
\cmidrule(lr){8-9} \cmidrule(lr){10-11} \cmidrule(lr){12-13}
& Dice & IoU
& Dice & IoU
& Dice & IoU
& Dice & IoU
& Dice & IoU
& Dice & IoU \\
\midrule
CLIP
& \underline{93.7} & 88.8
& \underline{69.0} & \underline{58.2}
& 91.3 & 86.2
& 74.6 & 61.3
& 92.3 & 87.7
& 85.1 & 77.5 \\
UniMed-CLIP
& \textbf{94.3} & \underline{89.7}
& 62.9 & 53.1
& \textbf{94.2} & \textbf{89.6}
& \underline{77.9} & \underline{65.1}
& \textbf{96.0} & \textbf{93.0}
& \underline{86.5} & \underline{79.8} \\
FetalCLIP
& 83.4 & 73.7
& 21.1 & 11.9
& 77.9 & 67.0
& 54.3 & 40.2
& 90.8 & 84.3
& 69.8 & 60.2 \\
\midrule
SonoCLIP(w/o)
& \textbf{94.3} & \textbf{90.0}
& \textbf{74.1} & \textbf{64.8}
& \underline{91.5} & \underline{86.5}
& \textbf{78.1} & \textbf{65.6}
& \underline{93.2} & \underline{89.6}
& \textbf{87.2} & \textbf{80.5} \\
\bottomrule
\end{tabular}%
}
\end{table}

\subsection{Segmentation on the FetalP5 Dataset}
Table~\ref{tab:five_plane_seg} demonstrates that SonoCLIP achieves the best overall segmentation performance, with a Dice score of 87.2\% and an mIoU of 80.5\%. The qualitative results in Fig.~\ref{fig:zero_shot__seg_open_data}(b) further demonstrate that the generated mask predictions are clearer and more complete. These findings indicate that the proposed pre-training strategy generates stronger dense visual representations, which effectively transfer to downstream tasks in fetal ultrasound segmentation.

\section{Conclusions}
\label{sec:con}
We present SonoCLIP, a vision–language foundation model for fetal ultrasound. By introducing a mask-channel visual pathway and a sigmoid pairwise contrastive loss, SonoCLIP learns region-aware representations robust to heterogeneous supervision and large-scale implicit negatives. Extensive experiments show consistent gains in cross-center zero-shot transfer and cross-dataset generalization, as well as strong performance under frozen-encoder classification and segmentation. These results indicate that SonoCLIP can serve as a transferable backbone for fetal ultrasound applications, including standardized plane recognition and anatomy-aware analysis, with potential to support reliable assessment in routine and resource-limited clinical settings.

\subsubsection{Acknowledgments.} This work was supported in part by the National Key Research and Development Program of China under Grant 2023YFC2705700, in part by the National Natural Science Foundation of China under Grants 62225113 and U23B2048, in part by the Innovative Research Group Project of Hubei Province under Grants 2024AFA017, in part by the Science and Technology Major Project of Hubei Province under Grants 2024BAB046 and 2025BCB026, and in part by the New Cornerstone Science Foundation through the XPLORER PRIZE. This work was also supported by WHU-Kingsoft Joint Lab. The numerical calculations in this paper have been done on the supercomputing system in the Supercomputing Center of Wuhan University.

\bibliographystyle{splncs04}
\bibliography{main}

\end{document}